\documentclass[10pt,twocolumn]{article}
\pdfoutput=1
\usepackage{simpleConference}
\usepackage{times}
\usepackage{graphicx}
\usepackage{amssymb}
\usepackage{url,hyperref}

\begin{document}

\title{Semi-Supervised Model Training for Unbounded Conversational Speech Recognition}
\author{Shane Walker, Morten Pedersen, Iroro Orife and Jason Flaks \\
Marchex Inc., 520 Pike, Seattle, WA, 98101 \\
}

\maketitle
\thispagestyle{empty}

\begin{abstract}
For conversational large-vocabulary continuous speech recognition (LVCSR) tasks, up to about two thousand hours of audio is commonly used to train state of the art models. Collection of labeled conversational audio however, is prohibitively expensive, laborious and error-prone. Furthermore, academic corpora like Fisher English (2004) or Switchboard (1992) are inadequate to train models with sufficient accuracy in the unbounded space of conversational speech. These corpora are also timeworn due to dated acoustic telephony features and the rapid advancement of colloquial vocabulary and idiomatic speech over the last decades.

   Utilizing the colossal scale of our unlabeled telephony dataset, we propose a technique to construct a modern, high quality conversational speech training corpus on the order of hundreds of millions of utterances (or tens of thousands of hours) for both acoustic and language model training. We describe the data collection, selection and training, evaluating the results of our updated speech recognition system on a test corpus of 7K manually transcribed utterances. We show relative word error rate (WER) reductions of \{35\%, 19\%\} on \{agent, caller\} utterances over our seed model and 5\% absolute WER improvements over IBM Watson STT on this conversational speech task. 
   \end{abstract}

\section{Introduction}

This paper examines a semi-supervised approach that aims to increase the quantity of conversational telephony speech transcripts available to train a LVCSR system. We define dataset construction and training as semi-supervised because we employ a \textit{seed} model to transcribe a vast quantity of unlabeled audio, perform data selection on the new transcripts, retrain the seed model and then repeat the process with the improved decoder \cite{li2016semi}. Our approach works by running a large-beam decoder tuned for high accuracy on our unlabeled telephony dataset. Lattices generated during the decoding process are used to compute Minimum Bayes Risk (MBR) confidences. The transcribed text is filtered to select minimum-length utterances with the lowest MBR confidence \cite{xu2011minimum} and the lowest language model (LM) perplexity. Perplexity is a measurement of how well a probabilistic LM will predict new sample text. We use an LM trained on 20K manually transcribed in-domain conversational utterances. 

This method takes advantage of the scale of Marchex's call traffic, enabling us to rapidly construct a very large-scale speech dataset, covering all types of language contexts, speaker demographics, accents and noise conditions. It also permits tracking of changes in quotidian vernacular as well as changes in acoustic channel features based on shifts in device and codec technology.

Because the error rate of the confidence-filtered training data can limit the gains due to poor acoustic modeling alignments \cite{vesely2013sequence}\cite{wang2007unsupervised}, we use various natural language processing (NLP) heuristics to algorithmically identify the highest prevalence, unique mistranscriptions for correction. We have developed tools to facilitate the creation and application of corrective text transforms over the full corpus of automatic transcriptions. The updated, post-processed text improves the quality of our acoustic model training alignments and so we iterate anew by retraining our acoustic model from scratch with the transformed text. 

For language modeling, the set of unique, text transform targets (i.e. the applied corrections) are added to the ground truth set used to build a new LM for the next iteration of filtering and utterance perplexity scoring. The core of our approach, an iterative method of taking qualified output from a seed model, using various NLP heuristics to further correct and select utterances for use in the subsequent rounds of training, allows us to progressively reduce the error rate of our ASR models, while operating at a scale that allows us to generalize well on in-domain speech.

The paper is organized as follows, Section 2 will provide some perspective on the complexity of the conversational ASR task. Section 3 will review recent schemes for speech dataset construction, especially for large-scale and low-resourced tasks. Section 4 describes our Speech Recognition system. Section 5 introduces the Marchex U.S. English Corpus of conversational North American english, discusses semi-supervised training and the data pipeline. Section 6 presents our results and contrasts our corpus with other conversational corpora. Section 7 describes how we scale up and areas of future work.

\section{Conversational ASR}

Automatic speech recognition (ASR) of spontaneous conversations is a different and more complex task than ASR for Voice Command or Voice Search applications performed by modern digital assistants. In addition to the usual challenges in LVCSR (e.g. speaker-independence, coarticulation, variable speech rates, noise-robustness and LM capacity) in natural, unscripted conversations additional factors come into play. These include \textit{disfluencies} such as mid-sentence hesitations, stutters, ungrammatical or filled pauses (uh, um, ah, er), back-channels (yeah, mhm, uh-huh), discourse markers (like, so, you know), self-editing terms (or rather, I mean), cut-off phrases, restarts, repetitions, final lengthening of syllables, coughs and laughter \cite{lease2006recognizing}\cite{deshmukh1998resegmentation}. 

In an article which tackles the linguistic appropriations and interpretations of Chomsky's ``Colorless Green Ideas Sleep Furiously", Manfred discusses a co-operative principle in human communication which binds two speakers to conversational maxims \cite{jahn2002colorless}. For speakers and listeners, this amounts to a set of interpretive assumptions that are very flexible in the presence of ungrammatical, rhetorical, figurative or completely novel utterances. This means that in free flowing conversation, semi-grammatical incongruities and semantic ill-formedness are always admissible when the utterance is well-chosen and/or the listener obtains a meaningful interpretation. Now considering that the English language has approximately half a million words, excluding many colloquial forms, with Unabridged English dictionaries listings of between 300,000 to 600,000 words, we see the combinatorial complexity and valid correct transcription of an arbitrary spontaneous utterance though finite, is unbounded \cite{Macquarie:99}.

\section{Building ASR Training Corpora}

There are many approaches to building speech training sets, including acoustic data perturbation and data synthesis \cite{ko2015audio}. Our survey of the literature will be restricted to unsupervised and semi-supervised approaches to corpora assembly. 

Google takes advantage of their large scale in constructing a training set for their Voice Search and Voice Input tasks for low-resource languages such as Brazilian Portuguese, Italian, Russian and French \cite{kapralova2014big}. Their unsupervised approach makes use of a slow but accurate decoder, confidence scores, transcript length and transcript flattening heuristics to select the utterances for acoustic modeling. 

In conjunction with owner-uploaded transcripts, Youtube apply ``island of confidence" filtering heuristics to generate additional semi-supervised training data for the deep neural network (DNN) based acoustic model (AM) driving their closed captions feature \cite{liao2013large}.

Kapralova \textit{et al}. and Yu \textit{et al}. \cite{kapralova2014big}\cite{yu2010unsupervised} train acoustic models on a Mandarin language Broadcast News (BN) and Broadcast Conversation (BC) dataset created with semi-supervised techniques. Due to the prevalence of English loan words and code-switching, data selection starts with a dual Mandarin-English language classifier, followed by the computation of utterance and word-level decoder confidence scores for the Mandarin-only utterances.

Ragni \textit{et al}. \cite{ragni2014data} use a semi-supervised system to build corpora for low-resource languages Zulu and Assamese task, using weighted word-confusion-network confidences for data selection. 

Li \textit{et al}. \cite{li2016semi} employ semi-supervised methods to construct a Mandarin training corpus based on a Chinese television spoken lectures series, using conditional random fields (CRF) for confidence estimation instead of the raw ASR decoder confidence measure. 

Enarvi \textit{et al}. \cite{enarvi2013studies}\cite{kurimo2015modeling} tackle a conversational Finnish language ASR task with a novel semi-supervised approach to training text selection.  In lieu of adding new transcribed candidate utterances to the corpus based on low in-domain LM perplexity, they score utterances by the \textit{decrease in} in-domain perplexity when the utterance is removed from the set of candidate utterances.

For a low-resource English, German and Spanish LVCSR task, Thomas \textit{et al}. \cite{thomas2013deep} use a hybrid confidence score based on word-level ASR confidence as well as a posteriogram-based phoneme occurrence confidence. This latter confidence uses a posteriogram representation of an utterance computed by passing utterance acoustic features through a trained DNN classifier.

\section{Speech Recognition System}

The seed ASR system is based on an online decoder written using Kaldi, a free, open-source C++ toolkit for speech recognition research \cite{povey2011kaldi}. In online decoding, the input audio features are processed buffer by buffer, progressively emitting the output text with minimal latency and without having to ingest the entire input before producing output. The seed decoder uses a ``prebuilt" deep neural network - hidden Markov model (DNN-HMM) hybrid model provided with Kaldi. In this hybrid model, a DNN is trained via minibatch asynchronous stochastic gradient descent to emit HMM posterior probabilities. These are then converted into ``scaled likelihoods" for the states of an HMM. In contrast to Gaussian mixture models (GMM) traditionally used in speech recognition, DNN models are superior classifiers that generalize better with a smaller number of model parameters, even when the dimensionality of the input features is very high \cite{vesely2013sequence}. Cross-entropy loss is the DNN training objective function, a standard choice for classification tasks.

The seed decoder's DNN is a 4 hidden layer neural network where the final layer is a softmax layer with a dimension corresponding to each of the 3500 context-dependent HMM states \cite{rath2013improved}. 

The input feature pipeline consumes 25 millisecond frames, processed to generate 13-dimensional Mel-frequency cepstral coefficients (MFCCs), which are spliced together with $\pm 3$ frames of context, for a total of $7 \cdot 13$ frames. The input dimensionality is then reduced to 40 by applying linear discriminant analysis (LDA) followed by a decorrelation step using maximum likelihood linear transform (MLLT). Finally a speaker normalization transform is applied, called feature-space maximum likelihood linear regression (fMLLR) \cite{povey2011kaldi}. 

During decoding the DNN takes an input feature vector 140 elements wide, comprising the 40 ``cooked" features described above \textit{and} a 100 element iVector. The same iVector is used for all acoustic feature vectors associated with a given speaker's utterances in the training set. Augmenting a new speaker's input feature vector with a corresponding iVector projection before DNN processing, permits the DNN to discriminate better between phonetic events in an adaptive, speaker-independent fashion \cite{gupta2014vector}, with minimal impact to the DNN training cycle.

The seed model was trained on 1935 hours of conversational audio. We extend it by \textit{rebuilding} its decoding-graph to incorporate an additional 20K manually transcribed in-domain language modeling utterances and expand the lexicon by an additional 600 domain-specific phonetic pronunciations. The lexicon is based on CMUdict, but with numeric stress markers removed.

The seed decoder's language model is a trigram model created from text from 1.6M (Fisher English) utterances using the SRILM toolkit. The lexicon, acoustic and language models are compiled down to weighted finite state transducers (WFSTs), which are composed into a single structure called the decoding-graph. Each letter stands for a separate WFST performing a specific input-to-output transduction: $HCLG = min(det(H \circ C \circ L \circ G))$.

    \begin{enumerate}
    \item $H$ maps multiple HMM states to context-dependent triphones.
    \item $C$ maps triphone sequences to monophones.
    \item $L$ is the lexicon, it maps monophone sequences to words.
    \item $G$ represents a language model FST converted from an ARPA format n-gram model.
    \end{enumerate}
     
    When the graph is \textit{composed} with an utterance's per-frame DNN output (i.e. HMM state likelihoods) it produces a lattice. The best path through the lattice produces text. For further details on ASR with weighted finite-state transducers refer to \cite{mohri2008speech}. To summarize, our seed ASR system is a prebuilt Kaldi \textit{online-nnet2}, cross-entropy trained, hybrid DNN-HMM model. It has an updated lexicon and language model and provides a competitive and well-understood baseline upon which we iterate.

\section{Training System Description}

First we introduce a brand new conversational telephony speech corpus of North American english and then describe our semi-supervised training and data selection methods in detail.

\subsection{The Marchex US English Corpus}

Marchex's call and speech analytics business securely fields over one million calls per business day, or decades of encrypted audio recordings per week. These are conversational, consumer to business phone calls occurring via a modern mixture of mobile phones on various telephone networks or landlines, capturing everyday North American dialog in every possible accent variant, English language fluency, under broad environmental or noise conditions, with comprehensive, colloquial vocabulary. Speaker demographics are extensive from teenagers to octogenarians. Example conversations may be sales related, e.g. calling to book a hotel, buying a mobile phone, cable service or to renegotiate insurance rates. Other examples are service related, e.g. scheduling a dentist appointment, an oil change, car repair or a house-moving service. The average conversation is four minutes long. Both the caller and the answering agent channels are recorded. This unlabeled corpus of calls is current, exhibiting natural and spontaneous conversations on business matters, in addition to popular culture, sports, politics and chitchat on uncontroversial topics like the weather.

\subsection{Data Collection and Processing}

\begin{figure}[h]
  \begin{center}
    \includegraphics[width=3.5in]{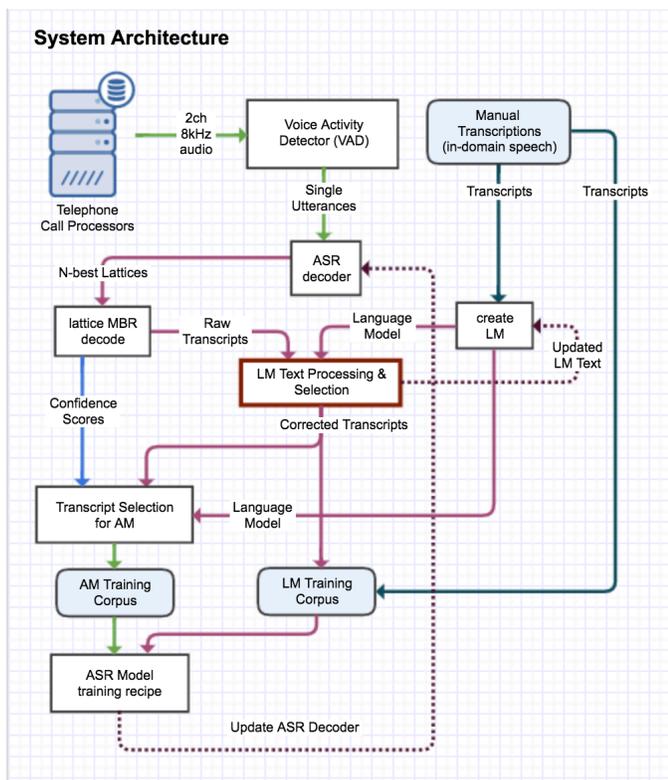}
  \end{center}
  \caption{\small Data Collection and Processing Flowchart}
\end{figure}

To make use of this telephony dataset, we programmatically gather call audio from the fleet of Marchex call processor servers. Mono 8kHz $\mu$-law-decoded files from the caller and agent channel are passed to a Voice Activity Detector (VAD) which creates single utterances usually shorter than 5 seconds long. For our initial experiments, we decoded a subset of 35 million utterances or some 25,000 hours of raw conversational audio with roughly a 44\%-56\% split between caller and agent. This split is less than even due to VAD's rejection of silent or degenerate caller-side audio, e.g. voicemail, fax machines calling phones.

The system architecture is shown in Figure 1. Solid lines show the flow of data towards the AM and LM training corpora. Dotted lines denote updates to the ASR decoder, as well as updates to language modeling text used for perplexity based data selection.

Similarly to \cite{kapralova2014big}, we decode using a slower but more accurate, non-production decoder tuned to have a large-beam. The decoder emits N-best lattices which we use to compute MBR confidences per utterance, with Kaldi's \textit{lattice-mbr-decode}. Our 20K manually transcribed corpus was similarly decoded and MBR scores were compared to Word Error Rate (WER). Shown below in Figure 2, a strong correlation between low MBR score and low WER suggests that MBR will useful for selecting accurately transcribed utterances. In the table below, we outline WER statistics for two values of very low MBR. Interestingly, the 90\%+ WER utterances turn out to either be systemic mistranscriptions or Spanish language IVR prompts.

\begin{center}
\begin{tabular}{|c|c|c|} \hline 
     \textbf{WER Statistics} & MBR=\textbf{0.0} & MBR=\textbf{0.1} \\ \hline
count & 337 & 1263 \\ \hline
mean & 4.14 & 20.3 \\ \hline
std & 0.0 & 0.0 \\ \hline
min & 0.0 & 0.0 \\ \hline
25\% & 0.0 & 0.0 \\ \hline
50\% & 0.0 & 0.0 \\ \hline
50\% & 0.0 & 0.0 \\ \hline
75\% & 0.0 & 0.0 \\ \hline
\textbf{90\%} & \textbf{6.25} & \textbf{28.5} \\ \hline
95\% & 20.0 & 50.0 \\ \hline
max & 100.0 & 100.0 \\ \hline
\end{tabular}	
\end{center}

Kaldi's \textit{lattice-confidence} is another confidence that we examined. Its value is the difference in total costs between the best and second-best paths through the N-best lattice. We ultimately rejected it due to lack of performance and low correlation with WER.

\begin{figure}[h]
  \begin{center}
    \includegraphics[width=3.5in]{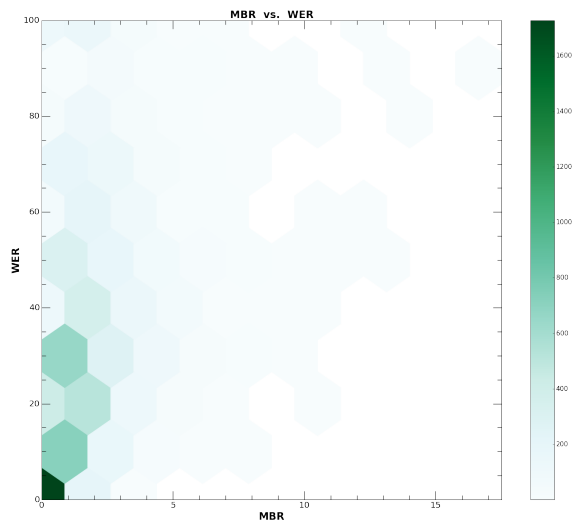}
  \end{center}
  \caption{\small Density plot of MBR versus WER}
\end{figure}

While MBR is used as a measure of expected risk in the ``whole system" based on the full N-best lattice, language model perplexity is a measurement of how well a probabilistic LM will predict a new sample text. Low perplexity indicates the LM is good at predicting the new text and is not ``confused". When combined, MBR and perplexity provide good intuition about how ``hard" it was for the system to arrive at its 1-best text output, the assertion being that lower WER utterances are ``easier" to decode. For selecting utterances, we compute perplexity with a Kneser-Ney smoothed 5-gram model with a 125K vocabulary and 5M n-grams \cite{chen1996empirical}.

Given the scale of our audio retrieval and the simplicity of the VAD used, there are a few different kinds of non-speech audio that get automatically transcribed that we certainly do not want in our training set. These include: hold-time musak, telephony signaling tones, Spanish language utterances (especially in IVR), pseudo random impulsive noises from typing on keyboards, cellphones dropping, Rihanna, laughter, coughing and other environmental noise. Utterances in the table below will have high perplexity, i.e. greater than 1000, when scored with a 5-gram LM trained on our 20K manually transcribed corpus. We remove up to 7M such degenerate utterances along with utterances with very short transcripts.

\begin{center}
\begin{tabular}{|l|l|} \hline
     \textbf{Removed Utterances} & \textbf{Audio content}   \\ \hline
	 be in they need to & ``bienvenidos" (Spanish) \\ \hline
	 but i spend you own & ``para espagnol" (Spanish) \\ \hline
	 bull pretty men dogs & ``oprima dos" (Spanish) \\ \hline
	 much guess seem go & ``marque cinqo" (Spanish) \\ \hline
	 it it it's it's it  & telephony signaling noise \\ \hline
	 whole whole whole &  telephonic beeps \\ \hline
	 mhm mhm mhm mhm & impulsive  noise \\ \hline
	 mm mm mm mm mm & impulsive noise \\ \hline
	 [noise] i i i &  environmental noise \\ \hline
	 and uh and uh and uh & hold music \\ \hline
	 or or or or or or & hold music \\ \hline
	 and uh and uh & Rihanna song \\ \hline
	 in a in a in in a & hold music \\ \hline  
\end{tabular}	
\end{center}

Next we look at other systemic errors that we can correct. The approach taken is based on global and sub-structure frequency of the \textit{seed} transcribed text. By sorting and counting the highest prevalence unique full utterances, we identify common elements where incomplete language representation and/or missing audio context can be fixed. Sub-structure frequencies are counted by using n-gram or part-of-speech tagging to isolate sub-elements to be amended. For example in the table below ``a grey day" can be part of ``you have a great day" or ``it is a great day".

\begin{center}
\begin{tabular}{|l|l|} \hline
     \textbf{Caller Mistranscriptions} & \textbf{Ground truth}  \\ \hline
have a grey day & have a great day \\ \hline
yeah that be great & yeah that'd be great \\ \hline
okay think so much & okay thanks so much \\ \hline
b. e. as in boy & b. as in boy \\ \hline
a two one zero & eight two one zero \\ \hline
i don't have any count & i don't have an account \\ \hline
\end{tabular}	
\end{center}

Identification and creation of targeted replacements are prepared manually via custom tools developed to present top candidates for correction. We distinguish caller versus agent side text because the nature of conversational speech on the caller side is much more diverse. Additionally, agent audio quality is usually higher, as agents may be in a call center or quiet office, while the caller may be in the car, on the street or on the bus. Below is a small section of agent side mistranscriptions, a number of which are from the Interactive Voice Response (IVR) utterances.

\begin{center}
\begin{tabular}{|l|l|} \hline
     \textbf{Agent Mistranscriptions} & \textbf{Ground truth} \\ \hline
horror leather increase & for all other inquiries (IVR) \\ \hline
rest you & press two (IVR) \\ \hline
oppressed you & or press two (IVR) \\ \hline
arrest three & press three (IVR) \\ \hline
um or cared & customer care \\ \hline
call back drone normal &  call back during normal \\ \hline
for parts in excess serious & for parts and accessories \\ \hline
active eight & activate \\ \hline
chevy taco & chevy tahoe \\ \hline
now and if you words & now in a few words \\ \hline
retire fritcher jack & free tire pressure check \\ \hline
\end{tabular}	
\end{center}

Once transforms have been created and applied to automatic utterances, the corrected text is ready to be filtered with both MBR and perplexity at thresholds appropriate for acoustic modeling. Language modeling text is also derived from the same corrected text but with much tighter perplexity thresholds, usually 40-80. From the original batch of 35M utterances, we are left with between 2.5M and 5M pristine utterances. We contrast these figures with the 1.6M utterances that comprise the Fisher English corpus \cite{cierifisher}.

\begin{figure}[h]
  \begin{center} 
    \includegraphics[width=3.5in]{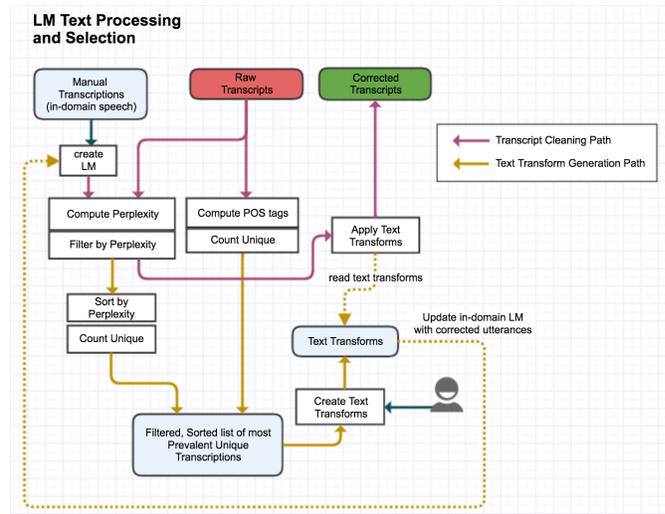}
  \end{center}
  \caption{\small Text Processing and Selection}
\end{figure}

Figure 3 details the \textit{Text Processing and Selection} block shown in Figure 1. We note two paths though the system. The first (solid magenta) is strictly for generating corrected text to build the AM and LM training corpora. The second path (yellow dotted) is for the generation of new LM building text derived transform targets (i.e. text corrections). These manual contributions are language modeling ground truth and are admissible to improve the capacity and the ability of our LM to generalize in subsequent training iterations. Starting with just the 20K manually transcribed corpus for language modeling, through this iterative process we grew the LM text used to compute perplexity in all parts of the system by another 6K items. This is a 30\% increase in the amount of high quality LM text and more importantly, text which comprises the correct labels for the most common in-domain conversational phrases.

\subsection{Retraining The ASR Model}

After we've successfully prepared the corrected, filtered automatic transcripts, it is time to retrain our ASR model. For retraining, we choose the cleanest 5,000 hours or 11.7M utterances. This figure was selected to have minimal impact on the training recipe's hyperparameters, with an eye out for maximum data-capacity of the training model. For AM training, we add in 13K utterances from our 20K manually transcribed set to the automatic, cleaned corpus. Because our manually transcribed utterances are the most accurate, these 13K are used for the training recipe's initial monophone and triphone steps. The remaining 7K utterances are excluded as a test set. Given the relative data increase over the seed model, we use a larger-capacity multi-splice version of the \textit{online-nnet2} recipe described in Section 4. This recipe uses 2 additional fully-connected hidden layers, for a total of 6, and a more elaborate input splicing scheme. We train this model on a single 12GB NVIDIA Titan X (Pascal) GPU for 6 full epochs over a period of 2 weeks.

\section{Experimental Results}

Our results are as follows on the Marchex North American english conversational task, a test set based on our 7K manually transcribed utterances, excluded in the training process described above. We compute WER scores for the seed model, IBM Watson Speech to Text service \cite{DBLP:journals/corr/SaonKRP15}\cite{DBLP:journals/corr/SaonSRK16} as well as our updated production model. Our updated model shows strong performance against IBM and demonstrates our ability to generalize well on an unseen dataset with a model trained on a mixture of manually transcribed and automatic transcriptions.

While IBM have not trained their models on Marchex English, their results are valid benchmarks because of the their published results on the Hub5 2000 Evaluation conversational task, a corpus of 40 \textit{test} telephone conversations from the SWBD and CALLHOME corpora \cite{DBLP:journals/corr/SaonKRP15}\cite{DBLP:journals/corr/SaonSRK16}. We contrast the proportions of Hub5 2000 with the size of our test set of 7K utterances or 4.5 hours of no-filler, manually transcribed conversational audio, sourced from more than 3,000 calls.

\begin{center}
\begin{tabular}{|l|c|c|} \hline
     Model & \textbf{Agent} WER & Relative gain\\ \hline
	 Seed model & 22.1 & - \\ \hline
	 IBM Watson STT & 20.0 & 9.5\% \\ \hline
	 Marchex production & \textbf{14.27} & \textbf{35\%} \\ \hline

\end{tabular}	
\end{center}

\begin{center}
\begin{tabular}{|l|c|c|} \hline
     Model &  \textbf{Caller} WER & Relative gain\\ \hline
	 Seed model & 21.6 & - \\ \hline
	 IBM Watson STT & 22.6 & -4.6\% \\ \hline
	 Marchex production & \textbf{17.5} & \textbf{19\%} \\ \hline

\end{tabular}	
\end{center}

\subsection{Comparing conversational corpora}

To better understand our performance with respect to the IBM models trained on Fisher English (FE) or Switchboard (SWBD), we now examine more closely how the Marchex English (ME) is different. By ME, we refer only to this first, \textit{post-seed} iteration of 11.7M utterances or 5,000 hours used to train an updated model. 

During the collection of FE, topics were pre-assigned or worked out between the contributors \cite{cierifisher}. For SWBD, a prompt suggested a topic of conversation \cite{godfrey1992switchboard}. ME on the other hand, captures real world conversations with the full degree of naturalness. FE furthermore excludes greetings and leave-takings, which we consider essential to correctly decode. SWBD transcribers were asked post-facto to rate the naturalness of conversations on a 5-point scale from ``very natural" to ``artificial-sounding". The mean rating for SWBD utterances is 1.48 \cite{godfrey1992switchboard}.

\begin{figure}[h]
  \begin{center}
    \includegraphics[width=3.5in]{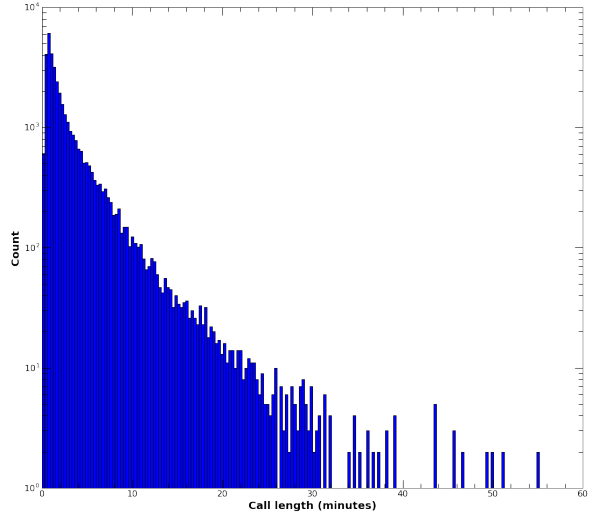}
  \end{center}
  \caption{\small Histogram of call lengths (from a sampling of 37K calls) }
\end{figure}

FE calls lasted no more than 10 minutes from which 8 minutes were deemed useful. ME calls last on average 4 minutes, but can be as short as 30 seconds, as in a voice-mail or wrong number. They can also be as long as an hour in the case that a contract is being negotiated or there are terms and conditions to be agreed upon. The much longer temporal context under which ME utterances are automatically collected, adds to the diversity of the corpus. In Figure 4 we show the distribution of durations in minutes of a 37K sampling of ME calls. In the table below, we draw further contrast between FE, SWBD and Marchex English \cite{deshmukh1998resegmentation}\cite{cierifisher}.

\begin{center}
\begin{tabular}{|l|c|c|c|} \hline
      & SWBD & FE & ME  \\ \hline
     Hours & 309 & 2,000 & 5,000+ \\ \hline
	 Speakers & 543 & 20,407 & 605K  \\ \hline
	 Utterances & 391K  & 1.6M & 11.7M \\ \hline
	 Conversations & 2,400 & 16,000+ & 288K \\ \hline
	 Words & 3M & 18M & 79.5M \\ \hline
\end{tabular}	
\end{center}

\section{Conclusions}

 In this report we have outlined results from only \textit{one} iteration of our semi-supervised approach. We review our plan to scale up and promising next steps.

\subsection{Scaling Up}

Encouraged by our very competitive error-rates, we see a lot of potential as our corpus grows. A natural question at this point, given an iterative process which produces increasingly large quantities of audio and text is: How do we scale up processing in a time and cost efficient manner?
 
 Given our goals of iterating on a monthly cadence, our solution is to use modern, cloud-based distributed computation. Initial work to collect, decode, clean utterances and train our models took place over a couple of months in a small local-cluster environment. So our first step was to move our corpus of 30M post-VAD utterances from the first round and well as 30M brand new utterances, hot off the wire, into an Amazon Web Services (AWS) S3 bucket. An S3 bucket is a logical unit of storage used to store data objects (audio and text) as well as any corresponding metadata like utterance ids, speaker-to-utterance mapping, etc.
  
 Next, we build an Amazon Machine Image (AMI), which provides the information required to launch a virtual machine instance, pre-configured with the requisite 64-bit system architecture, operating system, Python environment, Kaldi decoder and other software dependencies. Now we can spin up a dynamic and configurable cluster of virtual machines for re-decoding and post-processing. 
 
 To reliably manage the scheduling and distribution of audio to be re-decoded and VMs ready to accept work, we use Amazon Simple Queue Service (Amazon SQS). This service offers a highly-scalable hosted queue for sending, storing and receiving messages and is designed to guarantee that messages are processed exactly once, in the exact order that they are sent, with limited throughput \cite{AWSSQS}. 
 
 Now with a SQS queue, a fleet of VMs and S3 data, we are ready to re-decode. First we start off by populating our SQS queue with work items. This is done by generating a S3 Inventory Report, which is an enumeration of the 60M (audio) data objects to re-decode. We initialize our SQS queue from this report. Then we simply turn on the fleet and our SQS queue distributes messages to it. Messages are simply locations in S3 of audio (utterances) to decode. Our fleet consists of ``spot instances", a flexible, cost-effective alternative VM provisioning solution especially for data analysis, batch and background processing jobs where applications can be interrupted. At any time during processing if a VM goes away or stops, the message will merely timeout and go back into the queue to be rescheduled for processing by another VM. With a fleet of 100 mixed class \{cc2.8xlarge, r4.8xlarge, x1.16xlarge, m4.16xlarge\} VM instances, we are able to re-decode 30M utterances in 20 hours.  
 
 \begin{figure}[h]
  \begin{center}
    \includegraphics[width=3.5in]{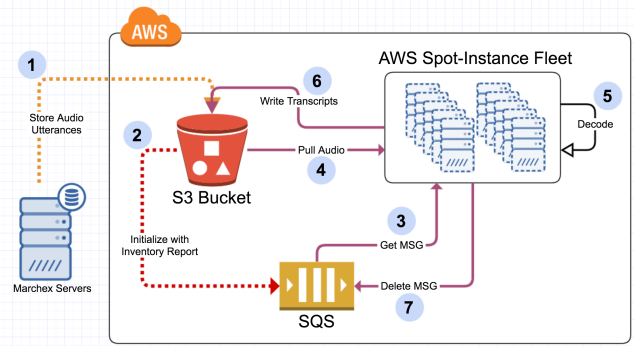}
  \end{center}
  \caption{\small Re-decoding AWS architecture}
\end{figure}

 Finally to do utterance filtering, processing and ASR model retraining, we employ a GPU-enabled AWS P2 instance like the p2.16xlarge. With 16 NVIDIA K80 GPUs each with 12 GB of memory, 64 virtual CPUs, 700+ GB of memory and low-latency, peer-to-peer GPU-to-GPU transfers, this class of machine is most commonly used for scientific and industrial scale deep learning tasks. While true distributed end-to-end ASR model training is an eventual objective, the P2 instance solution is most compatible with the parallelization tools in our Kaldi-based training recipe and provides immediate performance gains. In lieu of the weeks it took to train the first iteration using a local GPU, our AWS solution completes in a matter of days.

\subsection{Future Work}

Our semi-supervised training process permits us to compile very large, high quality conversational speech datasets orders of magnitude greater than what is possible via manual transcription. The manual effort is highly focused on specific tasks that have the highest impact on WER reduction and that improve the compounding effect of rinsing and repeating with a bigger and better decoder, trained on cleaner and larger quantities of correctly labeled audio. 

Future work includes making the VAD more selective, improving language detection and speech signal conditioning. There are also opportunities to use RNN-LM or CNN models for text classification to do more powerful data selection \cite{lai2015recurrent}. Furthermore, we see a lot of potential in the algorithmic superiority of bleeding-edge ASR methods using attention-based models or sequence trained neural networks with lattice-free MMI or CTC objectives \cite{amodei2015deep}\cite{povey2016purely}.

\bibliographystyle{abbrv}
\bibliography{refs}

\begin{thebibliography}{10}

\bibitem{AWSSQS}
{AWS Documentation, Amazon Simple Queue Service, Developer Guide}.
\newblock
  \url{http://docs.aws.amazon.com/AWSSimpleQueueService/latest/SQSDeveloperGuide/Welcome.html}.

\bibitem{amodei2015deep}
D.~Amodei, R.~Anubhai, E.~Battenberg, C.~Case, J.~Casper, B.~Catanzaro,
  J.~Chen, M.~Chrzanowski, A.~Coates, G.~Diamos, et~al.
\newblock Deep speech 2: End-to-end speech recognition in english and mandarin.
\newblock {\em arXiv preprint arXiv:1512.02595}, 2015.

\bibitem{chen1996empirical}
S.~F. Chen and J.~Goodman.
\newblock An empirical study of smoothing techniques for language modeling.
\newblock In {\em Proceedings of the 34th annual meeting on Association for
  Computational Linguistics}, pages 310--318. Association for Computational
  Linguistics, 1996.

\bibitem{cierifisher}
C.~Cieri, D.~Miller, and K.~Walker.
\newblock The fisher corpus: a resource for the next generations of
  speech-to-text.

\bibitem{deshmukh1998resegmentation}
N.~Deshmukh, A.~Ganapathiraju, A.~Gleeson, J.~Hamaker, and J.~Picone.
\newblock Resegmentation of switchboard.
\newblock In {\em ICSLP}. Syndey, 1998.

\bibitem{enarvi2013studies}
S.~Enarvi and M.~Kurimo.
\newblock Studies on training text selection for conversational finnish
  language modeling.
\newblock In {\em Proceedings of the 10th International Workshop on Spoken
  Language Translation (IWSLT 2013)}, pages 256--263, 2013.

\bibitem{godfrey1992switchboard}
J.~J. Godfrey, E.~C. Holliman, and J.~McDaniel.
\newblock Switchboard: Telephone speech corpus for research and development.
\newblock In {\em Acoustics, Speech, and Signal Processing, 1992. ICASSP-92.,
  1992 IEEE International Conference on}, volume~1, pages 517--520. IEEE, 1992.

\bibitem{gupta2014vector}
V.~Gupta, P.~Kenny, P.~Ouellet, and T.~Stafylakis.
\newblock I-vector-based speaker adaptation of deep neural networks for french
  broadcast audio transcription.
\newblock In {\em Acoustics, Speech and Signal Processing (ICASSP), 2014 IEEE
  International Conference on}, pages 6334--6338. IEEE, 2014.

\bibitem{jahn2002colorless}
M.~Jahn.
\newblock ‘colorless green ideas sleep furiously: A linguistic test case and
  its appropriations.
\newblock {\em Literature and Linguistics: Approaches, Models and Applications:
  Studies in Honour of Jon Erickson. Ed. Marion Gymnich, Ansgar N{\"u}nning and
  Vera N{\"u}nning. Trier: Wissenschaftlicher Verlag Trier}, pages 47--60,
  2002.

\bibitem{kapralova2014big}
O.~Kapralova, J.~Alex, E.~Weinstein, P.~J. Moreno, and O.~Siohan.
\newblock A big data approach to acoustic model training corpus selection.
\newblock In {\em INTERSPEECH}, pages 2083--2087, 2014.

\bibitem{ko2015audio}
T.~Ko, V.~Peddinti, D.~Povey, and S.~Khudanpur.
\newblock Audio augmentation for speech recognition.
\newblock In {\em INTERSPEECH}, pages 3586--3589, 2015.

\bibitem{kurimo2015modeling}
M.~Kurimo, S.~Enarvi, O.~Tilk, M.~Varjokallio, A.~Mansikkaniemi, and
  T.~Alum{\"a}e.
\newblock Modeling under-resourced languages for speech recognition. lang.
\newblock {\em Res. Eval}, pages 1--27, 2015.

\bibitem{lai2015recurrent}
S.~Lai, L.~Xu, K.~Liu, and J.~Zhao.
\newblock Recurrent convolutional neural networks for text classification.
\newblock In {\em AAAI}, volume 333, pages 2267--2273, 2015.

\bibitem{lease2006recognizing}
M.~Lease, M.~Johnson, and E.~Charniak.
\newblock Recognizing disfluencies in conversational speech.
\newblock {\em IEEE Transactions on Audio, Speech, and Language Processing},
  14(5):1566--1573, 2006.

\bibitem{li2016semi}
S.~Li, Y.~Akita, and T.~Kawahara.
\newblock Semi-supervised acoustic model training by discriminative data
  selection from multiple asr systems' hypotheses.
\newblock {\em IEEE/ACM Transactions on Audio, Speech and Language Processing
  (TASLP)}, 24(9):1520--1530, 2016.

\bibitem{liao2013large}
H.~Liao, E.~McDermott, and A.~Senior.
\newblock Large scale deep neural network acoustic modeling with
  semi-supervised training data for youtube video transcription.
\newblock In {\em Automatic Speech Recognition and Understanding (ASRU), 2013
  IEEE Workshop on}, pages 368--373. IEEE, 2013.

\bibitem{Macquarie:99}
R.~Mannell.
\newblock Infinite number of sentences.
\newblock Technical report, Macquarie University, Department of Linguistics,
  1999.

\bibitem{mohri2008speech}
M.~Mohri, F.~Pereira, and M.~Riley.
\newblock Speech recognition with weighted finite-state transducers.
\newblock In {\em Springer Handbook of Speech Processing}, pages 559--584.
  Springer, 2008.

\bibitem{povey2011kaldi}
D.~Povey, A.~Ghoshal, G.~Boulianne, L.~Burget, O.~Glembek, N.~Goel,
  M.~Hannemann, P.~Motlicek, Y.~Qian, P.~Schwarz, et~al.
\newblock The kaldi speech recognition toolkit.
\newblock In {\em IEEE 2011 workshop on automatic speech recognition and
  understanding}, number EPFL-CONF-192584. IEEE Signal Processing Society,
  2011.

\bibitem{povey2016purely}
D.~Povey, V.~Peddinti, D.~Galvez, P.~Ghahrmani, V.~Manohar, X.~Na, Y.~Wang, and
  S.~Khudanpur.
\newblock Purely sequence-trained neural networks for asr based on lattice-free
  mmi.
\newblock {\em Submitted to Interspeech}, 2016.

\bibitem{ragni2014data}
A.~Ragni, K.~M. Knill, S.~P. Rath, and M.~J. Gales.
\newblock Data augmentation for low resource languages.
\newblock In {\em INTERSPEECH}, pages 810--814, 2014.

\bibitem{rath2013improved}
S.~P. Rath, D.~Povey, K.~Vesel{\`y}, and J.~Cernock{\`y}.
\newblock Improved feature processing for deep neural networks.
\newblock In {\em Interspeech}, pages 109--113, 2013.

\bibitem{DBLP:journals/corr/SaonKRP15}
G.~Saon, H.~J. Kuo, S.~J. Rennie, and M.~Picheny.
\newblock The {IBM} 2015 english conversational telephone speech recognition
  system.
\newblock {\em CoRR}, abs/1505.05899, 2015.

\bibitem{DBLP:journals/corr/SaonSRK16}
G.~Saon, T.~Sercu, S.~J. Rennie, and H.~J. Kuo.
\newblock The {IBM} 2016 english conversational telephone speech recognition
  system.
\newblock {\em CoRR}, abs/1604.08242, 2016.

\bibitem{thomas2013deep}
S.~Thomas, M.~L. Seltzer, K.~Church, and H.~Hermansky.
\newblock Deep neural network features and semi-supervised training for low
  resource speech recognition.
\newblock In {\em Acoustics, Speech and Signal Processing (ICASSP), 2013 IEEE
  International Conference on}, pages 6704--6708. IEEE, 2013.

\bibitem{vesely2013sequence}
K.~Vesel{\`y}, A.~Ghoshal, L.~Burget, and D.~Povey.
\newblock Sequence-discriminative training of deep neural networks.
\newblock In {\em Interspeech}, pages 2345--2349, 2013.

\bibitem{wang2007unsupervised}
L.~Wang, M.~J. Gales, and P.~C. Woodland.
\newblock Unsupervised training for mandarin broadcast news and conversation
  transcription.
\newblock In {\em Acoustics, Speech and Signal Processing, 2007. ICASSP 2007.
  IEEE International Conference on}, volume~4, pages IV--353. IEEE, 2007.

\bibitem{xu2011minimum}
H.~Xu, D.~Povey, L.~Mangu, and J.~Zhu.
\newblock Minimum bayes risk decoding and system combination based on a
  recursion for edit distance.
\newblock {\em Computer Speech \& Language}, 25(4):802--828, 2011.

\bibitem{yu2010unsupervised}
K.~Yu, M.~Gales, L.~Wang, and P.~C. Woodland.
\newblock Unsupervised training and directed manual transcription for lvcsr.
\newblock {\em Speech Communication}, 52(7):652--663, 2010.

\end{thebibliography}
\end{document}